\documentclass[runningheads]{llncs}
\usepackage{graphicx}
\usepackage{subfigure}

\usepackage{tikz}
\usepackage{comment}
\usepackage{amsmath,amssymb} 
\usepackage{color}

\usepackage{amsmath}
\usepackage{amssymb}
\usepackage{booktabs}
\usepackage{multirow}
\usepackage{makecell}

\usepackage[capitalize]{cleveref}
\crefname{section}{Sec.}{Secs.}
\Crefname{section}{Section}{Sections}
\Crefname{table}{Table}{Tables}
\crefname{table}{Tab.}{Tabs.}

\newcommand{\tabincell}[2]{\begin{tabular}{@{}#1@{}}#2\end{tabular}}
\newcommand{\etal}{{\emph{et al.}}}

\usepackage[accsupp]{axessibility}  

\usepackage[width=122mm,left=12mm,paperwidth=146mm,height=193mm,top=12mm,paperheight=217mm]{geometry}

\begin{document}
\pagestyle{headings}
\mainmatter
\def\ECCVSubNumber{2710}  

\title{OPAL: Occlusion Pattern Aware Loss for Unsupervised Light Field Disparity Estimation} 

\titlerunning{OPAL} 
\authorrunning{Peng Li, Jiayin Zhao et al.} 
\author{\vspace{10pt}Peng Li$^{\ast}$ \hspace{0.3em} Jiayin Zhao$^{\ast}$ \hspace{0.3em} Jingyao Wu \hspace{0.3em} Chao Deng \hspace{0.3em} $^\dag$Haoqian Wang \hspace{0.3em} $^\dag$Tao Yu}
\institute{$^\ddag$ Tsinghua University}

\maketitle

\begin{abstract}
Light field disparity estimation is an essential task in computer vision with various applications. 
Although supervised learning-based methods have achieved both higher accuracy and efficiency than traditional optimization-based methods, the dependency on ground-truth disparity for training limits the overall generalization performance not to say for real-world scenarios where the ground-truth disparity is hard to capture. 
In this paper, we argue that unsupervised methods can achieve comparable accuracy, but, more importantly, much higher generalization capacity and efficiency than supervised methods. 
Specifically, we present the \textbf{O}cclusion \textbf{P}attern \textbf{A}ware \textbf{L}oss, named OPAL, which successfully extracts and encodes the general occlusion patterns inherent in the light field for loss calculation. OPAL enables: i) accurate and robust estimation by effectively handling occlusions without using any ground-truth information for training and ii) much efficient performance by significantly reducing the network parameters required for accurate inference. 
Besides, a transformer-based network and a refinement module are proposed for achieving even more accurate results. 
Extensive experiments demonstrate our method not only significantly improves the accuracy compared with the SOTA unsupervised methods, but also possesses strong generalization capacity, even for real-world data, compared with supervised methods.
Our code will be made publicly available.
\end{abstract}

\section{Introduction}
\label{sec:intro}
\let\thefootnote\relax\footnotetext{$^{\ast}$Equal Contribution}
\let\thefootnote\relax\footnotetext{$^\dag$Corresponding Author}
\let\thefootnote\relax\footnotetext{$^\ddag$BNRist and Graduate School at ShenZhen, Tsinghua University}

Light field disparity estimation is a hot topic in computer vision with strong potential for being used in various 3D applications~\cite{huang2015light,ng2005light,Manuel2018Fundamentals}. 
    \begin{figure}
        \centering
        \includegraphics[width=0.5\textwidth,height=5cm]{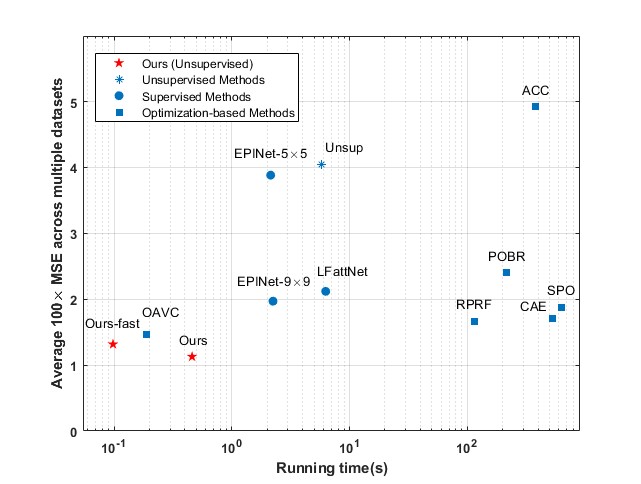}
        \vspace{-0.3cm}
        \caption{Comparisons of efficiency (running time) and performance (MSE$\times100$) with other state-of-the-art methods on 4D Light Field Benchmark~\cite{hci} and HCI Blender~\cite{hciold}. }
        \vspace{-10pt}
        \label{performance}
    \end{figure}
In contrast to stereo cameras~\cite{lee2000novel}, light field (LF) cameras  provide dense sub-aperture images (SAIs), making disparity estimation more accurate and robust. Moreover, compared with sensors like structured light~\cite{structurelight} and TOF cameras~\cite{tof}, LF cameras does not require active lighting and is less susceptible to strong infrared light interference, which makes them also available in outdoor scenarios.

Recently, many traditional~\cite{han2021novel,acc,CAE,ZhangSPO,occ} and learning-based approaches\\ \cite{heber2016,shi,epinet,lfatt,huang} have been proposed for disparity (or depth) estimation from light field images.
Traditional optimization-based methods could provide relatively accurate disparity results.
However, the time-consuming iterative optimization process or the relatively case-specific hyper parameters (for achieving best accuracy) limit the overall performance. 
Although supervised learning-based methods provide a good balance between accuracy and efficiency. Nonetheless, the heavy dependency on the dataset deteriorates the generalization capacity of such methods~\cite{unsup}, especially for real-world light field images. 

Among all the light field disparity estimation methods, unsupervised methods eliminate the prerequisite of ground-truth disparity for network training, thus having the potential for general, accurate and efficient disparity estimation as demonstrated in recent works~\cite{unsup,Zero,2020TIP}.
However, existing unsupervised methods still suffer from occlusions and tend to produce even larger disparity estimation errors around those regions. The reason is that they use the photometric consistency loss as the “only cue” for disparity estimation. Note that one of the most related works to ours is~\cite{han2021novel}, in which it uses an explicit voting strategy as a post-processing step for achieving occlusion aware disparity estimation. 
However, it doesn't consider the inherent patterns of the occlusion for disparity estimation, thus exceedingly limiting the final estimation accuracy. 

The key observation of our method is that, despite photometric consistency, the inherent occlusion pattern in light fields is another crucial cue for unsupervised learning. 
And if we can make the network “sense” the inherent occlusion patterns during the training process, we can eliminate the post-processing (e.g., post-voting) steps in~\cite{han2021novel} and achieve a fully end-to-end unsupervised framework that is not only more accurate but also more efficient and robust.
To fulfill this goal, we propose OPAL: \textbf{O}cclusion \textbf{P}attern \textbf{A}ware \textbf{L}oss, which incorporates occlusion patterns for calculating the photometric consistency loss directly. 
Specifically, for a pixel in the reference view, to calculate its OPAL, we first analyze and approximate its 2D occlusion pattern using a fixed number of 1D occlusion patterns in several directions. 
Note that a 1D occlusion pattern can further be approximated by a little number of pre-defined patterns based on the observation that occlusions in light fields mainly occur from a single side in most cases. After getting the occlusion pattern, we can use it to mask out the occluded views when calculating the photometric consistency loss. 
Note that OPAL is only used for training, and the trained model successfully learned how to adaptively select the optimal occlusion patterns for more accurate disparity estimation.
To further improve the accuracy, we propose a transformer-based encoder~\cite{transformer} for integrating features of epipolar plane images (EPIs) with high inter-view similarity (named EPI-Transformer) and a refinement strategy based on SAIs gradient maps.

As shown in~\cref{performance} and~\cref{runtime_table}, our network achieves the best overall accuracy cross different datasets even when compared with the SOTA supervised methods, which demonstrates the strong generalization capacity of our method. We also conduct mesh rendering comparisons (\cref{point_cloud}) to clearly demonstrate the superior performance of our method on real-world datasets.
Meanwhile, we only need significantly less time ($6$ hours V.S. a week of \cite{lfatt}) for network training and the run-time efficiency meets the SOTA standard. 

The main contributions can be concluded as:
\begin{itemize}
    \item[$\bullet$] We propose an \textbf{O}cclusion \textbf{P}attern Awar\textbf{E} \textbf{Net}work (namely OPENet) for unsupervised light field disparity estimation, which is not only much more accurate and efficient than current unsupervised methods, but also more general than existing supervised methods, 
    \item[$\bullet$] We present \textbf{O}cclusion \textbf{P}attern \textbf{A}ware \textbf{L}oss (\textbf{OPAL}), which enables strong occlusion handling ability for accurate and robust light field disparity estimation in an End-to-End and unsupervised manner, 
    \item[$\bullet$] We design an EPI-Transformer based on the Transformer encoder and a refinement module to further improve the overall performance of the OPENet.
\end{itemize}

\section{Related Work}
Recently, deep learning methods with supervised learning-based have been widely used in LF disparity estimation due to their efficiency and excellent performance. The typical supervised methods are based on EPIs. Shin \etal~\cite{epinet} presented a multi-branch CNN, which uses epipolar geometry information by extracting the features of EPIs from four directions and merges these features to improve accuracy. After that,  Tsai \etal~\cite{lfatt} proposed to take all SAIs as input to build a cost volume for regularization and then regress the disparity by taking a weighted sum. Different from EPI-based methods, Shi \etal~\cite{shi} adopted FlowNet2.0 for LF disparity estimation and refine the initial depth map according to warping errors. To reduce computational overhead and running time caused by 3D CNN, Huang \etal~\cite{huang} proposed a lightweight network based on multi-disparity scale cost aggregation and introduced a boundary guidance subnetwork to improve performance. 

However, it is unavailable to the ground-truth disparity in the real world and significant performance degradation would occur when applying supervised models trained on synthetic datasets to real-world scenes.

For the first time, Peng \etal~\cite{unsup} put forward an unsupervised learning-based method that uses the constraints between SAIs for training. Furthermore, Zhou \etal~\cite{2020TIP} designed three unsupervised loss functions (defocus loss, photometric loss, and symmetry loss)  according to the inherent depth cues and geometry constraints of light field images. However, these two methods both performed poorly in occluded and textureless areas, which demonstrates that it is a tricky and crucial problem to handle occlusion without real disparity as supervision. Jin \etal~\cite{jin2021occlusionaware} proposed to produce several initial depth maps, then fuse them based on reliability maps and an occlusion mask (generated by a threshold) to obtain final depth. Due to artificially set hard threshold, decision errors would occur at the edge of occluded areas, resulting in massive estimation errors. On the other hand, the occlusion problem is always the key to optimization-based methods\cite{hciold,bolles1987epipolar,tao2013depth,acc,CAE,han2021novel}. Williem \etal~\cite{CAE} designed an angle entropy measurement and an adaptive defocus response function to construct data cost, but it cannot be practicable due to time-consuming global optimization. Taking advantage of GPU accelerating, Han \etal~\cite{han2021novel} proposed a fast and occlusion-aware method that separates refocused pixels by use of a vote threshold and utilizes the number of the separated pixels to select the correct depth. Although it achieves competitive performance to learning-based methods, the hyper parameters may need to be adjusted carefully when applying to different scenarios. More importantly, as shown in~\cref{realworld} and \cref{point_cloud}, it cannot achieve consistently well performance for complex real-world scenarios.

Meanwhile, attention-based architectures have been applied in various computer vision tasks, such as image recognition\cite{VIT}, object detection\cite{detr}, and stereo depth estimation\cite{sttr}. The Transformer\cite{transformer} is a non-recurrent structure that relies on the attention mechanism to merge the information of all tokens and model the global dependencies of input and output, which is of particular importance for the work presented here. Hence we adopt Transformer Encoder to fuse the information of EPIs for better performance.

\begin{figure*}[t]
    \centering
    \includegraphics[width=\linewidth]{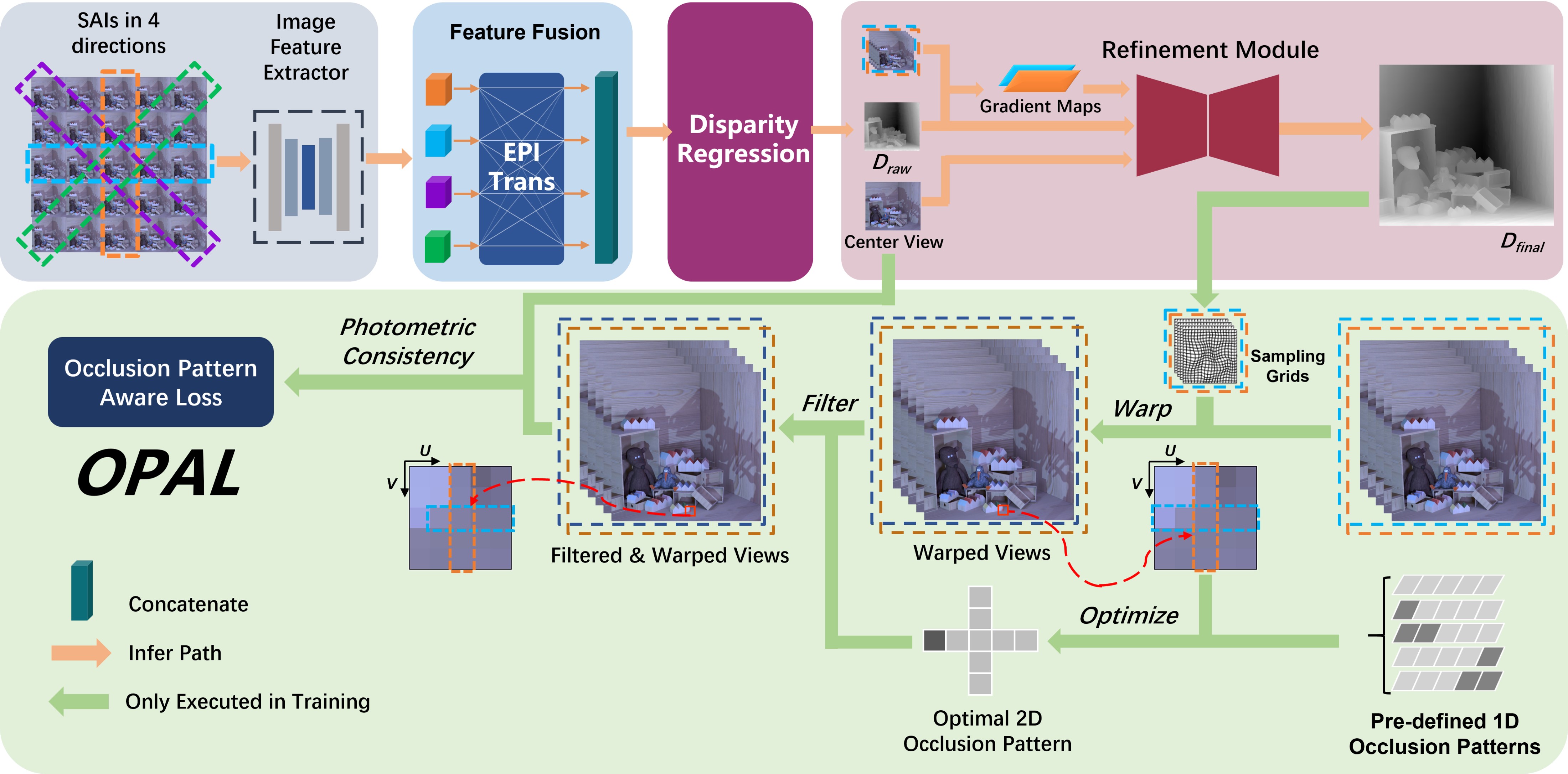}
    \vspace{-0.7cm}
    \caption{Overview of the proposed OPENet. It contains four components: feature extractor, EPI-Transformer, disparity regression and refinement module. The green part is the illustration of OPAL, which is only executed in training.}
    \label{network}
\end{figure*}

\section{Methodology}
The proposed OPENet contains feature extractor, EPI-Transformer, disparity regression and refinement module as illustrated in \cref{network}.  
The inputs of OPENet are SAIs from four angular directions (horizontal, vertical, left \& right diagonal). 
We use an SPP module~\cite{he2015spatial} with shared weights to extract global features of each view, which could better exploit the hierarchical contextual information and the correlation between adjacent regions. 
After getting the feature maps of different views, an EPI-Transformer is introduced to construct cross-view correlations for effective feature fusion. 
Then, the fused feature maps are fed into the disparity regression module to estimate the disparity of the center view. 
As in \cite{lfatt}, we obtain raw disparity $D_{raw}$ by regressing the probabilities of possible disparity and calculating the weighted sum of them,
 \begin{equation}
    D_{raw}=\sum\limits_{d =  - {D_{max}}}^{{D_{max}}} {d \times \sigma ( {{c_d}} )} \label{disp_regression}
\end{equation}\\
where $c_{d}$ is the cost for each possible disparity, $\sigma(\cdot)$ is softmax function and $D_{max}$ is the preset maximum disparity($D_{max}$ is set to 4 in this paper).
Finally, a gradient-based refinement module is introduced to better align disparity map and center input, and the output is denoted as $D_{final}$. 
Note that resolving the impact of occlusions for disparity estimation is a very challenging task not to say in an unsupervised  manner. We propose OPAL for training, which not only produces strong occlusion handling capacity but also  significantly improve the training and inference efficiency.
 



\subsection{Occlusion pattern aware loss}

Given a light field image ${L(x,y,u,v)} \in R^{HW \times N \times N \times 3}$, where $(x,y)$ and $(u,v)$ represent the 2D spatial and angular coordinate, and the resolution along spatial and angular are $H\times W$ and $N \times N$, respectively. Next, we use $x$ to represent $(x,y)$ and $u$ to represent $(u, v)$ for simplicity. 
Under the assumption of Lambertian and no-occlusion, the warped SAIs based on ground truth disparity should be the same as the central-view-image~\cite{jonschkowski2020matters},
\begin{equation}
    {I_{u_0}}(x) = {I_u} (x + (u-u_0)D(x)) \label{I_u}
\end{equation}
where ${D}$ is the ground truth disparity and $u_0$ is the angular coordinate of the central view. 
Based on \cref{I_u}, we can optimize the network by minimizing the photometric consistency loss~\cite{jonschkowski2020matters}. 
Theoretically, the assumption of photometric consistency is invalid in occluded areas since it mistakenly records the light reflected by other points. Our purpose is to teach the network to automatically filter out the occluded views for more accurate disparity estimation in an unsupervised manner.

\begin{figure}
    \centering
    \subfigure[]{
    \begin{minipage}[t]{0.3\linewidth}
    \centering
    \includegraphics[width=0.8\linewidth, height=0.8\linewidth]{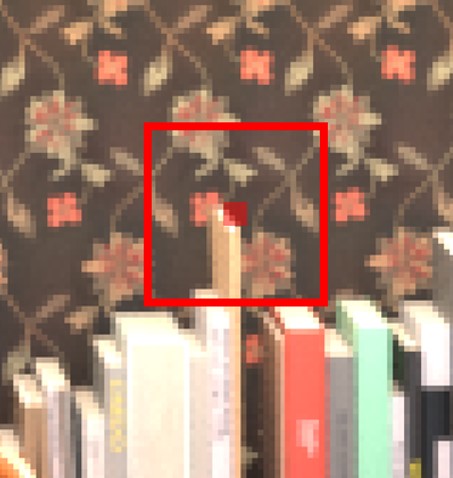}
    \label{}
    \end{minipage}
    }
    \subfigure[]{
    \begin{minipage}[t]{0.3\linewidth}
    \centering
    \includegraphics[width=0.8\linewidth]{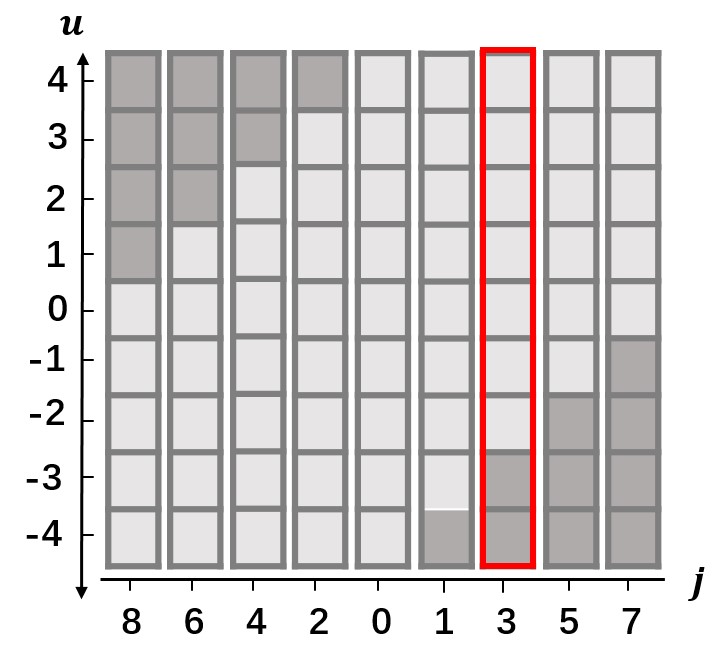}
    \label{filter}
    \end{minipage}
    }
    \subfigure[]{
    \begin{minipage}[t]{0.3\linewidth}
    \centering
    \includegraphics[width=0.8\linewidth, height=0.8\linewidth]{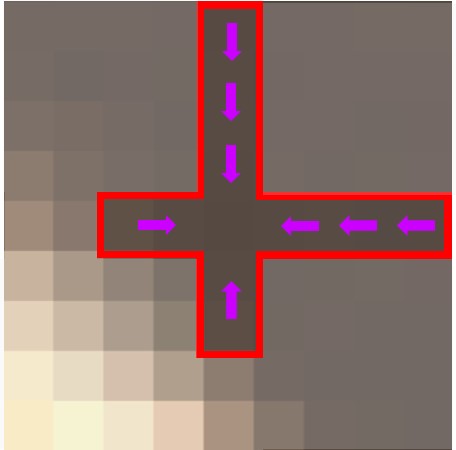}
    \label{illu_opal}
    \end{minipage}
    }
    \caption{Illustration of using occlusion pattern to handle occlusion issue. In (a), the red dot are occluded pixel from central view. (b) lists  all possible occlusion patterns in line when $N=9,\beta=1$, the red box represents the $OPL^\ast$ of red dot in (a). In (c), the red box means the $OPL^\ast$ to block the light from other scene points in horizontal and vertical directions.}
\label{fig}
\end{figure}

The most straightforward manner for achieving OPAL is to search the target occlusion pattern among all the possible occlusion patterns on the 2D angular plane. However, it is not trivial since there are massive possible occlusion patterns for each pixel.
Therefore, in OPAL, we approximate them as a collection of 1D patterns in several directions  and further simplify the 1D patterns by naturally hypothesizing: i) occlusion usually starts from one boundary view, ii) the occluded views must be neighbors. 
Specifically, we design $N $ proposal patterns in each direction for $N \times N $ LF images, denoted as $OPL$, as illustrated in \cref{filter}. 
For the vertical direction in Fig.\ref{illu_opal}, the top SAIs are not occluded, and the $OPL$ is defined by  
\begin{equation}
\label{opl_up}
    OP{L_j}(u) = \{ 
    \begin{array}{l} 
        1, u \ge -\frac{N - 1}{2} +\lceil {\frac{j}{2}} \rceil \\ 
        0,otherwise
    \end{array} 
\end{equation}
where $u$ is the angular index ranging from $-\frac{N-1}{2}$ to $\frac{N-1}{2}$, $j$ is the odd index of the occlusion pattern, and $\lceil{\cdot}\rceil$ is the ceil function. 
Conversely, if bottom views are not occluded, $OPL$ could be defined by 
\begin{equation}
\label{opl_down}
    OP{L_j}(u) = \{ 
    \begin{array}{l}
        1, u\le\frac{N - 1}{2} - \frac{j}{2}\\
        0,otherwise
    \end{array} 
\end{equation}
where $j$ is the even index.
Note that the quantity of possible $OPL$s will increase with $N$, which make it tough for the network to converge. 
For the datasets with high angular resolution, we down-sample the SAIs by a factor of $\beta$ to maintain stable performance.

During training, we first obtain photometric cost by warping the SAIs to the central view based on $\tilde{D}$ ($D_{raw}$ or $D_{final}$) from network,
\begin{equation}
    Cost_{opl}(j) = \frac{\sum_{u}[OPL_j(u)|\hat{I}_{u\to{u_0}}(x,\tilde{D})-I_{u_0}(x)|]}
    {\sum_{u}OPL_j(u)}
\end{equation}
where ${{\hat I_{u \to {u_0}}}( {x,\tilde D} )}$ represents the warped image based on output disparity. Afterward, we select the optimal occlusion pattern $OP{L^*}$ by minimizing the cost. Note that our strategy for generating $OPL^\ast$ is greedy, which may cause oscillations in non-occlusion areas. We set a threshold $\tau$ to determine if occlusion occurs. 
For non-occluded areas, we choose $OPL_0$ as $OPL^\ast$. 
Therefore $OPL^*$ could be defined as,
\begin{equation}
\label{opl}
    OPL^* = \{
    \begin{array}{l}
        OPL_0,\bigg| {Cost_{opl}(N)-Cost_{opl}(N-1)} \bigg| < \tau \\
        \mathop {\arg \min }\limits_{OPL} {Cost_{opl}(j)}, otherwise
    \end{array}
\end{equation}
where ${\tau}$ is set to 0.01 empirically. 
After that, we upsample $OPL{^*}$ by $\beta$ to meet the original light field angular resolution,
\begin{equation}
\label{upsample}
    OPLF = upsample(OPL^*, \beta)
\end{equation}

Due to the high correlation between $OPL^\ast$ and $\tilde{D}$, our OPENet could implicitly learn how to select $OPL^\ast$ after training. 
Eventually, the occlusion pattern aware loss in each direction is defined as:
\begin{equation}
\begin{aligned}
     &{\ell_{opal}}(\tilde{D}) =
    &{\sum\limits_{u}}{\sum\limits_x}{OPLF\cdot \big|{{{\hat I}_{u \to {u_0}}}( {x,\tilde D} ) - {I_{{u_0}}}( x )}\big|} \label{weight_rec}
\end{aligned}
\end{equation}\\
and the sum of $\ell_{opal}$ in all directions is used to penalize $\tilde{D}$.

\subsection{EPI-Transformer}
\label{transformaer}
Inspired by the attention mechanism and transformer-based methods in~\cite{transformer,VIT}, we propose EPI-Transformer to enhance cross-view correlations for more effective feature fusion. The input of EPI-Transformer is a sequence of angular tokens so that the information among SAIs could be fused. 
Considering that EPI-based methods utilize the proportional relationship between disparity and slope of EPIs and most pixels outside the focal plane are not aligned, we concatenate the feature of 2*2 neighboring pixels to compensate for this “misalignment”. Specifically, our EPI-Transformer consists of four identical fusion streams corresponding to the four directional inputs. For each stream, we first concatenate $N$ feature maps derived from the feature extractor and rearrange it from $R^{N \times HW \times C}$ to $R^{N \times HW/4 \times 4C}$, denoted as $E$, where $H \times W$ is the spatial resolution of SAIs and $4C$ is the embedding dimension.
Then we calculate position encoding according to \cref{PE}: 
\begin{equation}
\begin{cases}
PE( {pos,2i}) = \sin( {\frac{{pos}}{{{{10000}^{\frac{{2i}}{c}}}}}})\\
PE( {pos,2i + 1}) = \cos( {\frac{{pos}}{{{{10000}^{\frac{{2i}}{c}}}}}})
\end{cases} \label{PE}
\end{equation}
where ${pos}$ is the position of SAI in the epipolar plane and ${i}$ is the channel index. 
We take the sum of ${PE}$ and $E$ as the input embedding and put it through layer normalization. Afterward, we multiply input embedding by ${W_Q}$, ${W_K}$ and ${W_V}$ to generate ${Q}$, ${K}$ and ${V}$ and apply multi-head self-attention (${MHSA}$) to learn the dependence among different views:
\begin{equation}
    {MHSA}( {{E_A}} ) = Concat( {{H_1},...,{H_h}} ){W^O}\label{MHSA}
\end{equation}
\begin{equation}
    {H_i} = softmax( {\frac{{{Q_i}{K_i}^T}}{{\sqrt {C/h} }}} ){V_i}\label{H_i}
\end{equation}
where ${h}$ is the number of head groups, ${W^O}$ is the output projection matrix. Then we feed the sum of ${MHSA}$ and ${E}$ into a residual structure composed of layer normalization and MLP to obtain ${E_{fusion}}$. We get the final fused feature maps after another reshaping operation and additional convolution layers (a 3D convolution layer with kernel size $N \times 3 \times 3$ and a 2D convolution layer). Finally, the features in four directions are concatenated to regress the raw disparity $D_{raw}$.

\subsection{Refinement module}
\label{refinement}
Although our OPAL helps the model converge better and fast by handling most occlusion, we observe some dilation still appears in the object edges when foreground and background have similar brightness. To solve this issue, we introduce a refinement module to align the disparity with the input center view. We first warp SAIs in each direction to the central view based on $D_{raw}$ and then calculate the gradient maps between adjacent views for each pixel. Apparently, the average gradient indicates the probability of being miscalculated. As shown in \cref{network}, we concatenate the central view SAI, the raw disparity and the average gradient map together as input and sequentially pass it through the module designed in~\cite{sttr} to obtain the refined disparity in different directions, and the final disparity $D_{final}$ is the simple average of the refined disparity maps from all directions.

\subsection{OPENet-fast}
\label{openetfast}
By eliminating the refinement module and decreasing the directions (only using horizontal and vertical directions) as inputs, we can significantly improve the efficiency without sacrificing much accuracy, and we call this version OPENet-fast. Moreover, we only use the additional convolution layers mentioned in \cref{transformaer} for feature fusion.
As shown in \cref{synth_cross}, such a simple structure could also significantly improve the overall performance compared to existing methods, which also demonstrate the strong benefit of OPAL. 

\subsection{Implementation Details}
We also apply the edge-aware smoothness loss \cite{jonschkowski2020matters} to further improve the smoothness of textureless areas and provide disparity with sharp edges, 
\begin{equation}
    {\ell _{smooth}}( \tilde{D},I_{u_0} ) = | {{\partial _x}\tilde D} |{e^{ - \gamma | {{\partial _x}{I_{{u_0}}}} |}} + | {{\partial _y}\tilde D} |{e^{ - \gamma | {{\partial _y}{I_{{u_0}}}} |}}\label{smooth}
\end{equation}
where the edge weight $\gamma$ is set to 150. The total loss function is given by:
\begin{equation}
    {\ell _{total}} = \lambda_1 {\ell _{opal}^{raw}} + ( {1 - \lambda_1}){\ell _{opal}^{final}} + \lambda_2{\ell _{smooth}}\label{total_loss}
\end{equation}
where $\ell_{opal}^{raw}$ and $\ell_{opal}^{final}$ are OPAL (based on L1 norm) of $D_{raw}$ and $D_{final}$, respectively.  $\lambda_1$ is set to 0.6 and 1.0 for OPENet and OPENet-fast, respectively.  $\lambda_2$ is equal to 0.3.


We randomly crop ${64\times64}$ RGB patches from the LF images for training and the data augmentation strategy in EPINet~\cite{epinet} was used to improve robustness. 
We use Adam optimizer~\cite{Adam} with ${\beta_1}$ = 0.9 and ${\beta_2}$ = 0.999 and the batch size is set to 16. The initial learning rate is 1e-3 and is decreased to 1e-6. It only takes 210 epochs for convergence, and the training time is 6 hours for OPENet and 4 hours for OPENet-fast on a single NVIDIA RTX 3080, respectively. 

\section{Experiments}

In this section, we first introduce the datasets and metrics used for evaluation. Then we report the quantitative and qualitative comparisons with state-of-the-art methods. Finally, we perform ablation studies to analyze different components of our method.

\subsection{Datasets and evaluation criteria}
Our method is evaluated on three synthetic datasets (4D Light field Benchmark~\cite{hci}, the HCI Blender~\cite{hciold} and the dataset published by~\cite{newdata}) with ground-truth disparity and two kinds of real-world datasets provided by Stanford Lytro LF Archive~\cite{stanforddata} and Kalantari \etal~\cite{ref41}.

\noindent\textbf{Synthetic LF datasets.} The 4D Light field Benchmark~\cite{hci} and HCI Blender~\cite{hciold} are the most widely used datasets for evaluating light field disparity estimation.
The spatial resolution of the SAIs is ${512\times512}$ and the angular resolution is ${9\times9}$. 
For all learning-based methods, we use 16 scenes in a subset of the 4D Light field Benchmark (Additional) for training and evaluate all the methods on 8  representative scenes in both \cite{hci} and \cite{hciold} for better evaluating the generalization capacity. 
Note that we further verify the effectiveness of our method on the relatively simple and textureless scenes in \cite{newdata} directly without any fine-tuning process.

\begin{table}\scriptsize
\centering
\caption{Quantitative comparisons with other state-of-the-art methods on the 4D Light Field Benchmark~\cite{hci} and HCI Blender~\cite{hciold}}
\label{synth_cross}.
\resizebox{\textwidth}{!}{
\begin{tabular}{ll|llll|llll|l}
\toprule[1pt]
    \multicolumn{2}{c|}{\multirow{2}{*}{Methods}}  & \multicolumn{4}{c|}{4D LF Benchmark [ \textbf{MSE$\times$100} / \textbf{BadPix(0.07)}]}& \multicolumn{4}{c|}{HCI Blender [ \textbf{MSE$\times$100} / \textbf{BadPix(0.07)}]} & \multicolumn{1}{c}{\multirow{2}{*}{\textbf{
    \tabincell{c}{Cross-Dataset\\Average}}}}
    \\
    \multicolumn{2}{c|}{} & \makecell[c]{Boxes} & \makecell[c]{Cotton} & \makecell[c]{Dino} & \makecell[c]{Sideboard}& \makecell[c]{Buddha} & \makecell[c]{Buddha2} & \makecell[c]{MonasRoom} & \makecell[c]{Papillon}& \multicolumn{1}{c}{} 
    \\ 
\midrule[0.1pt]
    \multicolumn{1}{c}{\multirow{4}{*}{
    \tabincell{c}{Optimization\\-based}}} &
    ACC~\cite{acc} & \makecell[c]{14.151 / 27.294} & \makecell[c]{10.083 / 9.182} &  \makecell[c]{1.311 / 22.358} & \makecell[c]{5.588 / 24.660} &
    \makecell[c]{1.419 / 10.113} & \makecell[c]{0.523 / 10.467} &  \makecell[c]{0.657 / 10.562} & \makecell[c]{5.691 / 22.514} & \makecell[c]{4.928 / 17.144} 
    \\
    \multicolumn{1}{c}{} &
    SPO~\cite{ZhangSPO} & \makecell[c]{9.113 / 15.060} & \makecell[c]{1.376 / 2.858} &\makecell[c]{0.410 / 2.767} & \makecell[c]{1.143 / 9.547} &  
    \makecell[c]{0.554 / 2.336} & \makecell[c]{1.046 / 14.191} &\makecell[c]{0.563 / 6.690} & \makecell[c]{0.772 / 25.525} & \makecell[c]{1.872 / 9.872} 
    \\
    \multicolumn{1}{c}{} &
    CAE~\cite{CAE} & \makecell[c]{8.327 / 21.623} & \makecell[c]{1.804 / 5.281}  &\makecell[c]{0.382 / 14.175} & \makecell[c]{0.960 / 17.627} & 
    \makecell[c]{0.643 / 3.240} & \makecell[c]{\textbf{0.342 / 6.051}}  &\makecell[c]{0.493 / 7.314} & \makecell[c]{0.642 / \textbf{7.861}}  & \makecell[c]{1.699 / 10.397}  
    \\
     \multicolumn{1}{c}{} &
    OAVC~\cite{han2021novel} & \makecell[c]{6.99 / 16.1} & \makecell[c]{0.60 / 2.55} & \makecell[c]{0.27 / 3.94}& \makecell[c]{1.05 / 12.4} &  \makecell[c]{0.36 / 1.78} & \makecell[c]{1.29 / 11.7} & \makecell[c]{0.44 / \textbf{6.01}}& \makecell[c]{0.84 / 14.4} & \makecell[c]{ 1.46 / 8.61 } 
    \\
\midrule[0.1pt]
    \multicolumn{1}{c}{\multirow{3}{*}{Supervised}} &
    EPINet-5$\times$5~\cite{epinet}&\makecell[c]{6.140 / 12.462} & \makecell[c]{\textbf{0.200} / 0.549} &\makecell[c]{0.163 / 1.242} & \makecell[c]{0.818 / 4.763} 
    &\makecell[c]{0.775 / 4.516} & \makecell[c]{4.884 / 46.022} &\makecell[c]{2.170 / 19.670} &\makecell[c]{16.121 / 48.952}  &\makecell[c]{3.884 / 17.272} 
    \\
    \multicolumn{1}{c}{} &
    EPINet-9$\times$9~\cite{epinet}& \makecell[c]{6.054 / 11.891}  &\makecell[c]{0.230 / 0.492}  & \makecell[c]{0.181 / 1.222} &\makecell[c]{0.794 / 4.660} &
    \makecell[c]{0.392 / \textbf{1.543}}  &\makecell[c]{0.634 / 34.772}  & \makecell[c]{1.338 / 10.770} &\makecell[c]{6.126 / 35.564} &\makecell[c]{1.969 / 12.614}
    \\
    \multicolumn{1}{c}{} &
    LFattNet~\cite{lfatt} & \makecell[c]{\textbf{4.090 / 9.841}}  & \makecell[c]{0.210 / \textbf{0.252}}  & \makecell[c]{\textbf{0.080 / 0.752}} &\makecell[c]{\textbf{0.502 / 2.591}} & 
    \makecell[c]{0.324 / 2.028}  & \makecell[c]{6.060 / 34.233}  & \makecell[c]{0.782 / 10.757} &\makecell[c]{4.900 / 34.812} &  \makecell[c]{2.164 / 11.909} 
    \\
\midrule[0.1pt]
     \multicolumn{1}{c}{\multirow{5}{*}{Unsupervised}} &
    Unsup~\cite{unsup}  & \makecell[c]{11.356 / 45.126} & \makecell[c]{6.464 / 30.179} & \makecell[c]{1.893 / 29.664}& \makecell[c]{4.550 / 26.893} &
    \makecell[c]{1.269 / 12.372} & \makecell[c]{1.362 / 30.894} & \makecell[c]{1.683 / 19.294}& \makecell[c]{3.724 / 27.933} & \makecell[c]{4.038 / 15.455} 
    \\
    \multicolumn{1}{c}{} &
    Mono~\cite{2020TIP}  & \makecell[c]{9.749 / 22.884} & \makecell[c]{1.081 / 3.808} & \makecell[c]{0.657 / 5.402}& \makecell[c]{2.795 / 10.947} &
    \makecell[c]{- / -} & \makecell[c]{- / -} & \makecell[c]{- / -}& \makecell[c]{- / -} & \makecell[c]{- / -} 
    \\
    \multicolumn{1}{c}{} &
    OccUnNet~\cite{jin2021occlusionaware}  & \makecell[c]{7.45 / 26.24} & \makecell[c]{0.80 / 8.46} & \makecell[c]{0.63 / 8.25}& \makecell[c]{1.79 / 14.20} &
    \makecell[c]{0.34 / 4.11} & \makecell[c]{- / -} & \makecell[c]{0.57 / 10.57}& \makecell[c]{1.11 / 36.36} & \makecell[c]{- / -} 
    \\
    \multicolumn{1}{c}{} &
    Ours fast& \makecell[c]{5.520 / 17.173} & \makecell[c]{0.546 / 2.331} & \makecell[c]{0.428 / 3.952}& \makecell[c]{1.141 / 9.691} & \makecell[c]{\textbf{0.312} / 2.819} & \makecell[c]{1.523 / 20.628 } & \makecell[c]{0.428 / 7.709}& \makecell[c]{0.626 / 18.94} & \makecell[c]{1.315 / 10.405} 
    \\
    \multicolumn{1}{c}{} &
    Ours & \makecell[c]{4.928 / 14.622} & \makecell[c]{0.431 / 1.728} & \makecell[c]{0.320 / 3.634}& \makecell[c]{1.010 / 8.426} &  \makecell[c]{0.323 / 2.322} & \makecell[c]{1.058 / 12.44} & \makecell[c]{\textbf{0.367} / 6.732}& \makecell[c]{\textbf{0.581} / 12.145} & \makecell[c]{\textbf{1.127} / \textbf{7.756}} 
    \\
\bottomrule[1pt]
\end{tabular} }
\end{table}

\noindent\textbf{Real-world LF datasets.} Stanford Lytro LF Archive dataset~\cite{stanforddata}. consists of 251 real-world scenes captured by Lytro Illum camera. 
The dataset published by Kalantari \etal~\cite{ref41} is split into 72 training scenes and 25 testing scenes.
The two real-world datasets have the same spatial and angular resolution, which are
${376\times541}$ and ${14\times14}$, respectively. Note that we only apply the central ${9\times9}$ SAIs for disparity estimation.  
Since the real-world datasets could not provide ground-truth disparity, we only retrain the unsupervised models using the training set in~\cite{ref41} and test all the methods on other scenes of two datasets. 


\begin{figure}[t]
  \centering
  \begin{subfigure}{\leftline{\scriptsize Cotton}}
    \includegraphics[width=12.2cm,height = 3cm]{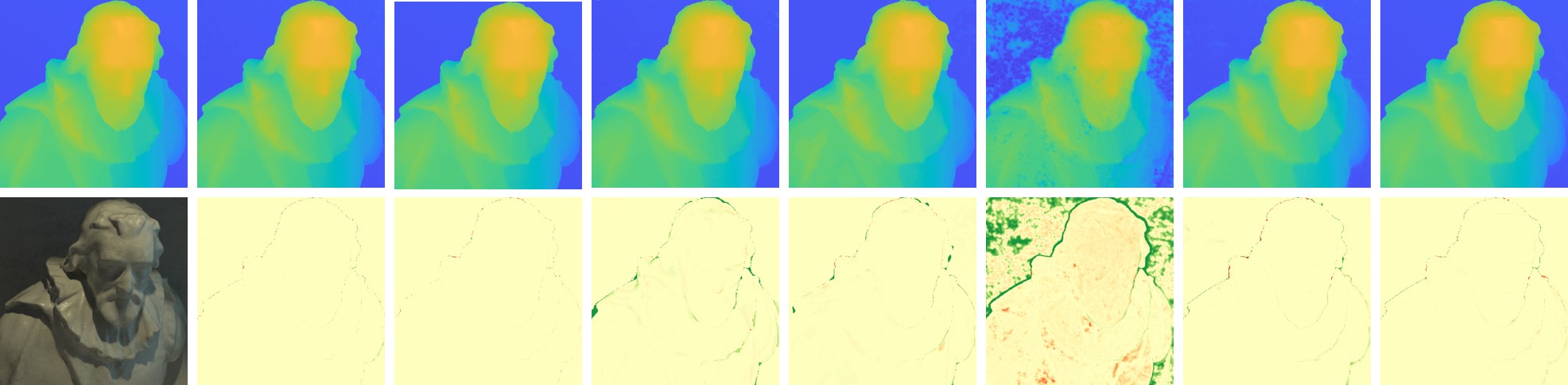}
  \end{subfigure}
  \begin{subfigure}{\leftline{\scriptsize  Sideboard}}
    \includegraphics[width=12.2cm,height = 3cm]{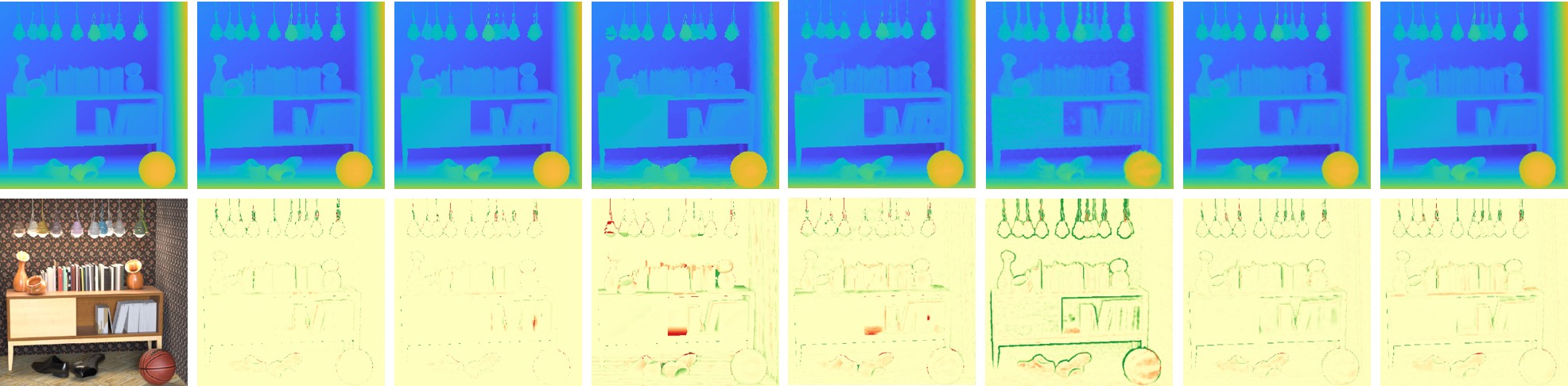}
  \end{subfigure}
  \begin{subfigure}{\leftline{\scriptsize  Buddha}}
    \includegraphics[width=12.2cm,height =3cm]{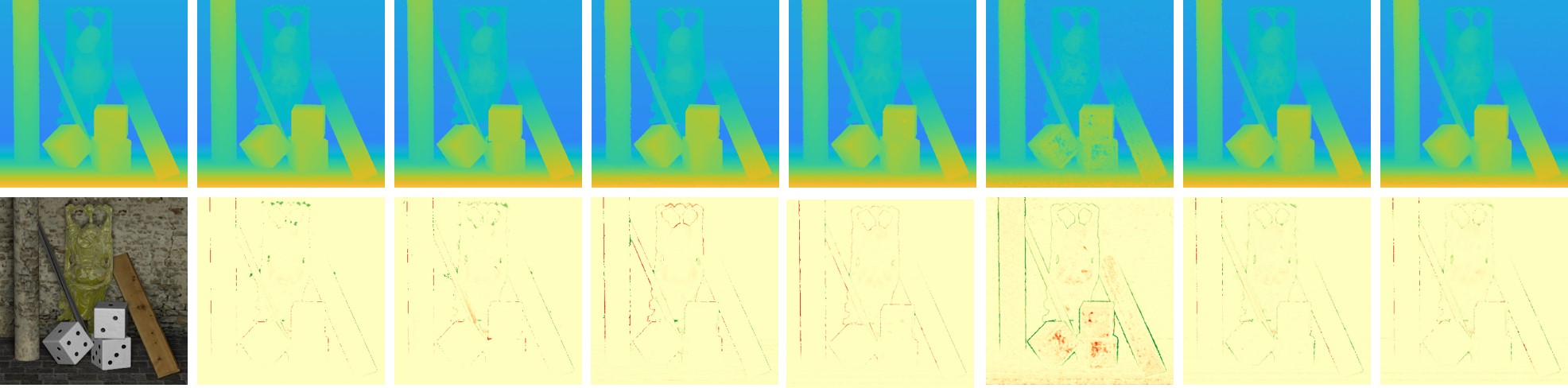}
  \end{subfigure}
  \begin{center}
    	\scriptsize \leftline{Ground Truth \quad(a)\qquad\qquad(b)\qquad\qquad~~(c)\quad\qquad~~(d)\qquad\qquad(e)\qquad\quad Ours fast~~~\quad~Ours}
  \end{center}  
  \vspace{-0.6cm}
\caption{Qualitative comparisons of the disparity and error maps on 4D Light field Benchmark~\cite{hci} and HCI Blender~\cite{hciold}. The compared methods include: (a)EPINet-9$\times$9~\cite{epinet}, (b)LFattNet~\cite{lfatt}, (c)CAE~\cite{CAE}, (d)OAVC~\cite{han2021novel} and (e)Unsup~\cite{unsup}.}
\label{hci}
\vspace{-0.6cm}
\end{figure}

\begin{figure}[t]
    \centering
    \includegraphics[width=0.5\textwidth]{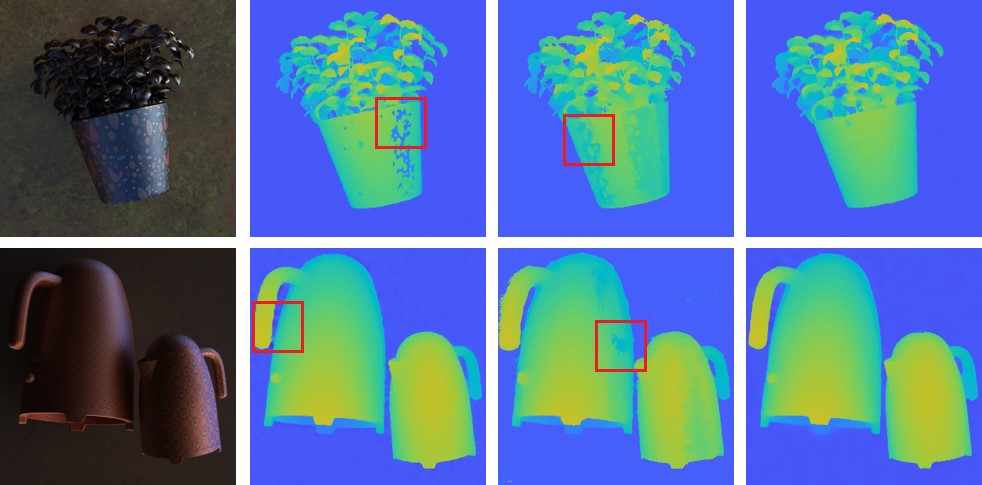}
    \begin{center}
    \vspace{-0.1cm}
    	\scriptsize \leftline{\qquad\qquad\qquad\qquad~~~~~ ~Central View EPINet-9$\times$9~~~~~SPO\qquad~~~~Ours}
  \end{center} 
  \vspace{-0.8cm}
  \caption{Qualitative Comparisons on dataset~\cite{newdata}.}
    \label{newdata}
\end{figure}

\begin{figure*}[t]
    \centering
    \includegraphics[width=\linewidth]{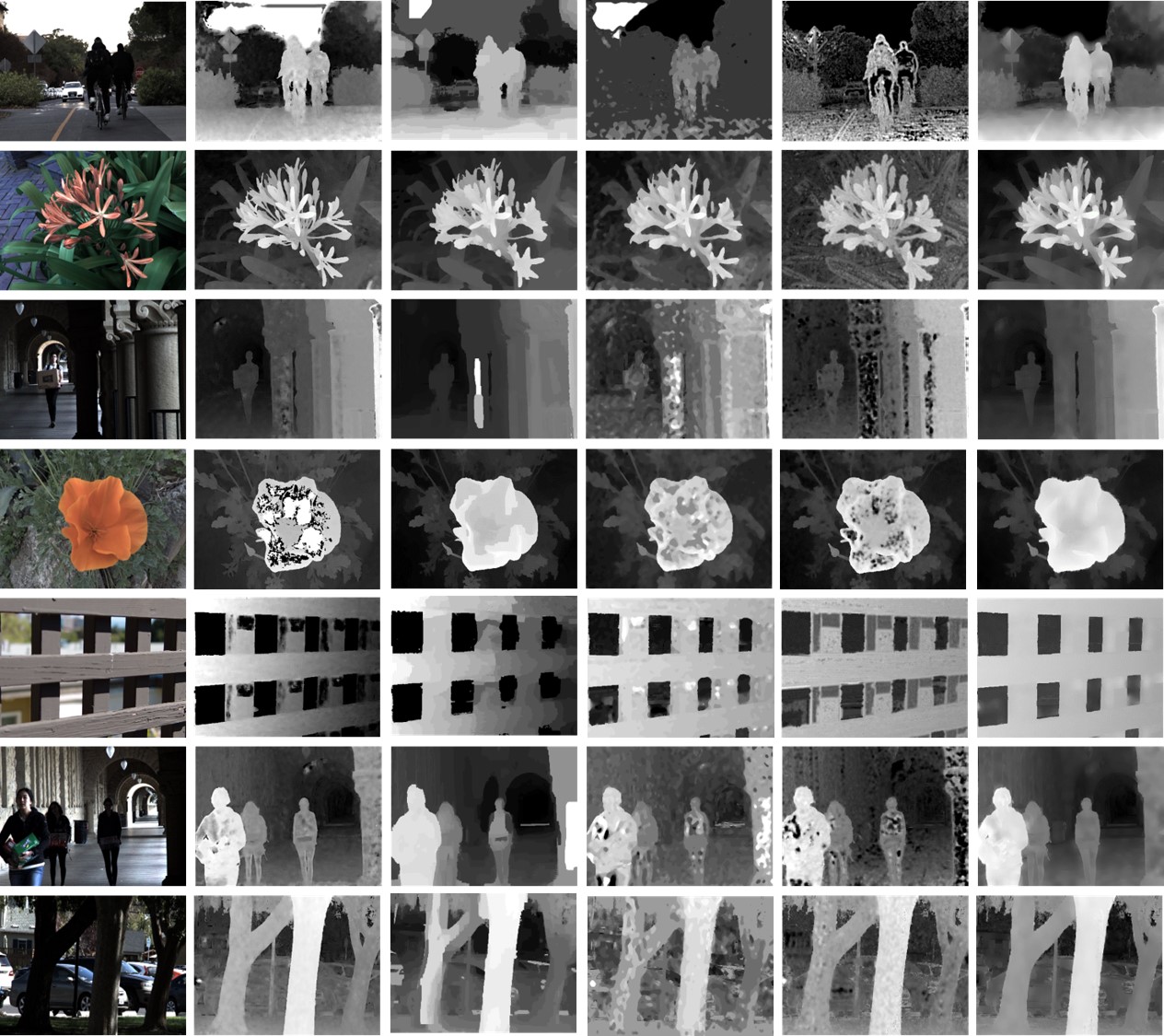}
     \begin{center}
     \vspace{-0.2cm}
    	\scriptsize \leftline{\qquad~Input~~~~ \qquad\qquad(a)~~\qquad\qquad\quad(b)~~\qquad\qquad\quad~~(c)\qquad\qquad\quad~~(d)~~\qquad\qquad~~~ours}
  \end{center}  
  \vspace{-0.8cm}
  \caption{Qualitative comparisons of the disparity maps estimated by different methods on real-world datasets~\cite{stanforddata,ref41}. The compared methods include: (a)LFattNet~\cite{lfatt}, (b)CAE~\cite{CAE}, (c)OAVC~\cite{han2021novel} and (d)Unsup~\cite{unsup}.}
  \label{realworld}
  \vspace{-0.4cm}
\end{figure*}

\begin{figure}[t]
    \centering
    \includegraphics[width=\linewidth,height =0.5\linewidth]{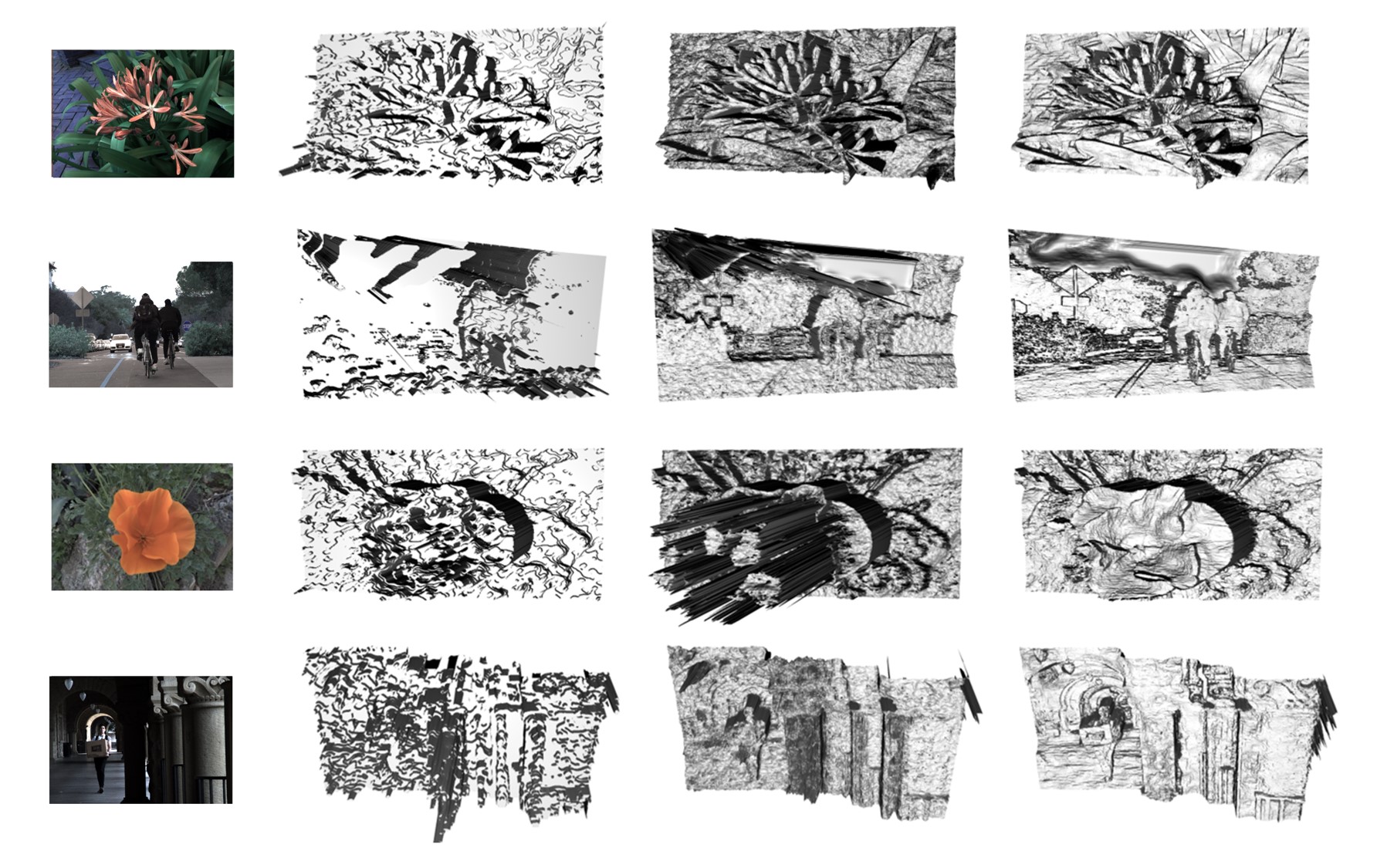}
     \begin{center}
     \vspace{-0.5cm}
    	\footnotesize \leftline{~~\qquad Input\qquad\quad\quad~~~OAVC~\cite{han2021novel}\qquad\qquad ~~~LFattNet~\cite{lfatt}\qquad\qquad~~~~~Ours}
  \end{center}  
  \vspace{-0.9cm}
  \caption{Mesh rendering comparisons on real-world datasets (\cite{stanforddata,ref41}) with SOTA optimization-based (OAVC~\cite{jin2021occlusionaware}) and supervised methods (LFattNet\cite{lfatt}).}
  \label{point_cloud}
  \vspace{-0.2cm}
\end{figure}

\begin{table}[t]\scriptsize
\centering
\caption{Comprehensive comparisons of efficiency via running time, parameters, MSE${\times100}$ and BadPix(0.07) on synthetic datasets}
\label{runtime_table}
\begin{tabular}{lllll}
\toprule[0.7pt]
    \multicolumn{1}{c}{Methods} & 
    \makecell[c]{Runtime/s} & 
    \makecell[c]{Parameters/M} &
    \makecell[c]{ MSE$\times$100} &
    \makecell[c]{ BadPix(0.07)}\\
\midrule[0.3pt]
    \makecell[c]{ACC~\cite{acc}} &
    \makecell[c]{628.36}  & 
    \makecell[c]{-} & 
    \makecell[c]{4.928} &
    \makecell[c]{17.144} \\
    \makecell[c]{SPO~\cite{ZhangSPO}} &
    \makecell[c]{380.64}  & 
    \makecell[c]{-} & 
    \makecell[c]{1.872} &
    \makecell[c]{9.872} \\
    \makecell[c]{CAE~\cite{CAE}} & 
    \makecell[c]{529.13} &  
    \makecell[c]{-} &      
    \makecell[c]{1.699} &
    \makecell[c]{10.397} \\
    \makecell[c]{OAVC~\cite{han2021novel}} & 
    \makecell[c]{0.19} &  
    \makecell[c]{-} &      
    \makecell[c]{1.46} &
    \makecell[c]{8.61} \\
\midrule[0.3pt]
    \makecell[c]{EPINet-5$\times$5~\cite{epinet}} & 
    \makecell[c]{2.14}    & 
    \makecell[c]{5.113}   & 
    \makecell[c]{3.884} &
    \makecell[c]{17.272} \\
    \makecell[c]{EPINet-9$\times$9~\cite{epinet}}&
    \makecell[c]{2.24}    &   
    \makecell[c]{5.118}  & 
    \makecell[c]{1.969} &
    \makecell[c]{12.614} \\
    \makecell[c]{LFattNet~\cite{lfatt}}&
    \makecell[c]{6.28}  & 
    \makecell[c]{5.058}  & 
    \makecell[c]{2.164} &
    \makecell[c]{11.909} \\
\midrule[0.3pt]
    \makecell[c]{Unsup~\cite{unsup}} & 
    \makecell[c]{5.73}  & 
    \makecell[c]{-}  &   
    \makecell[c]{4.038} &
    \makecell[c]{15.455}\\
    \makecell[c]{Ours fast} & 
    \makecell[c]{\textbf{0.098}}  & 
    \makecell[c]{\textbf{0.774}} & 
    \makecell[c]{1.315} &
    \makecell[c]{10.405}\\
    \makecell[c]{Ours} &
    \makecell[c]{0.46} &      
    \makecell[c]{1.047 } & 
    \makecell[c]{\textbf{1.127}} &
    \makecell[c]{\textbf{7.756}}\\
\bottomrule[0.7pt]
\end{tabular}
\end{table}

\begin{table}[t]\scriptsize
\centering
\caption{Ablation study on $OPAL$, down-sampling factor $\beta$, loss items and our network's componments on 4D Light Field Benchmark~\cite{hci}}
\label{abla_tab}
\begin{tabular}{l|lllll}
\toprule[1pt]
\multicolumn{1}{c}{\multirow{2}{*}{Methods}}&\multicolumn{5}{c}{MSE$\times$100}\\
\multicolumn{1}{c}{} & \makecell[c]{Boxes} & \makecell[c]{Cotton} & \makecell[c]{Dino} & Sideboard & \makecell[c]{Average}\\
\midrule[0.5pt]
Ours fast w/o $OPAL$              & \makecell[c]{8.368} & \makecell[c]{2.834} & \makecell[c]{1.027}& \makecell[c]{3.084} & \makecell[c]{3.828} \\
Ours fast ($\beta=1$)& \makecell[c]{5.562} & \makecell[c]{0.841} & \makecell[c]{0.509} & \makecell[c]{\textbf{1.000}} & \makecell[c]{1.978} \\
Ours fast ($\beta=2$) & \makecell[c]{5.467} & \makecell[c]{1.109} & \makecell[c]{0.509} & \makecell[c]{1.225} & \makecell[c]{2.007} \\
Ours fast ($\beta=4$) & \makecell[c]{5.520} & \makecell[c]{0.546} & \makecell[c]{0.428} & \makecell[c]{1.141} & \makecell[c]{1.908} \\
\midrule[0.5pt]
Ours w/o refinement & \makecell[c]{5.297} & \makecell[c]{0.510} & \makecell[c]{0.394} & \makecell[c]{1.080} & \makecell[c]{1.820} 
\\
Ours w/o EPI-Trans       & \makecell[c]{5.371} & \makecell[c]{0.462} & \makecell[c]{0.341} & \makecell[c]{1.06} & \makecell[c]{1.808}
\\
\midrule[0.5pt]
Ours w/o $\ell_{D_{raw}}$ & 
\makecell[c]{5.133} & 
\makecell[c]{0.431} & 
\makecell[c]{0.335} & 
\makecell[c]{1.12} & 
\makecell[c]{1.755}
\\
Ours w/o $\ell_{smooth}$ & 
\makecell[c]{6.775} & 
\makecell[c]{0.990} & 
\makecell[c]{0.612} & 
\makecell[c]{1.553} & 
\makecell[c]{2.482} 
\\
\midrule[0.5pt]
Ours & \makecell[c]{\textbf{4.928}} & \makecell[c]{\textbf{0.431}} & \makecell[c]{\textbf{0.320}} & \makecell[c]{1.010} & \makecell[c]{\textbf{1.672}}
\\
\bottomrule[1pt]
\end{tabular}
\end{table}

\noindent\textbf{Metrics.} We use  Bad Pixel Ratio (BPR) defined in~\cite{hci} and Mean Square Errors (MSE) for quantitative evaluation on synthetic datasets. Note that BPR measures the percentage of mistakenly estimated pixels whose errors exceed ${\varepsilon}$, which is set to 0.07 for evaluation.

\subsection{Comparisons with state-of-the-art methods}
\noindent\textbf{Methods for comparison.} We quantitatively and qualitatively compare our methods with other state-of-the-art methods, including optimization-based methods (ACC~\cite{acc}, CAE~\cite{CAE}, SPO~\cite{ZhangSPO}, OAVC~\cite{han2021novel}), supervised learning-based methods (EPINet-$5\times5$, EPINet-$9\times9$~\cite{epinet}, LFattNet~\cite{lfatt}) and existing unsupervised methods (Unsup~\cite{unsup},Mono~\cite{2020TIP},OccUnNet~\cite{jin2021occlusionaware}). Due to the code of Mono~\cite{2020TIP} and OccUnNet~\cite{jin2021occlusionaware} is unavailable, we only show some results in their papers. 

\noindent\textbf{Quantitative comparison on synthetic data.} For synthetic datasets, we train the model with simplified occlusion patterns (${\beta = 4}$) for better performance. \cref{synth_cross} reports the results of all methods. \cref{hci} visualizes the disparity and error maps of different methods.
For the dataset of~\cite{newdata}, \cref{newdata} shows the results of some better methods in two scenes.
\begin{itemize}
    \item[$\bullet$]OPENet-fast achieves comparable performance to the optimization-based methods on the accuracy, while OPENet surpasses them significantly. For areas with poor texture or severe occlusion, our method perform better.
    \item[$\bullet$]Compared to the previous unsupervised methods~\cite{unsup,2020TIP,jin2021occlusionaware}, ours could handle most occlusion cases well and generate global smooth disparity, thus greatly surpassing them. 
    \item[$\bullet$]Supervised methods achieve better performance on 4D Light field Benchmark~\cite{hci} than ours. However, significant performance degradation occurs when generalizing them to other datasets (right in \cref{synth_cross} and \cref{newdata}). The performance of our method is much robust across different datasets. 
\end{itemize}

\noindent\textbf{Qualitative comparisons on real-world results.} \cref{realworld} shows the visual results of different methods. The disparity maps generated by the previous unsupervised method~\cite{unsup} have obvious errors in occluded and texture-less areas. 
Due to severe noise and complex structures in real-world scenarios, even the SOTA optimization-based approach (OAVC) cannot maintain good performance with carefully adjusted parameters.
And the supervised methods inevitably suffer serious performance degradation. By comparison, our method could generate sharper and cleaner disparity on real-world LF images. In \cref{point_cloud}, we convert the disparity of some cases into 3D point clouds using the same parameters. The impressive results show our method tends to provide more accurate disparity, even could be used for further 3D applications. 

\noindent\textbf{Comprehensive comparisons of efficiency and generalization.} As listed in \cref{runtime_table}, we comprehensively compare the efficiency and accuracy of other state-of-the-art methods and ours via running time, parameters, MSE${\times100}$ and BadPix on synthetic datasets. 
As shown in \cref{runtime_table}, the overall accuracy of our method outperforms SOTA methods. Although LFattNet achieves the best accuracy in \cite{hci} (\cref{synth_cross}), the performance decreases significantly in \cite{hciold} due to different data modalities. Note that the performance of CAE also decreases in \cite{hci} although it performs the best in \cite{hciold}. This obviously demonstrates the strong generalization capacity of our unsupervised method. Moreover, we would like to mention the high run-time efficiency  (0.46s V.S. 6.28s[LFattNet]) and training efficiency (6 hours V.S. 1 week[LFattNet]) of our method. Note that by fully taking advantage of GPU acceleration, OAVC runs faster than OPENet but still cannot surpass OPENet-fast on both efficiency and accuracy. This is because OPAL is strong enough to guide the training to achieve fast convergence, and we can use a lightweight model (1.047Mb) for achieving high efficiency without sacrificing  accuracy. To ensure fairness, all the methods are tested on an NVIDIA RTX 3080 GPU PC. 

\subsection{Ablation study}
We first perform ablation studies on 4D Light field Benchmark~\cite{hci} to evaluate the effectiveness of the proposed $OPAL$. Specifically, we train our OPENet-fast without $OPLF$ and with three kinds of $OPLF$ ($\beta=1,2,4$), respectively. As shown in \cref{abla_tab}, the model trained without $OPLF$ performs much worse. In addition, the network trained with non-simplified $OPLF$ ($\beta=1$) would perform better in severely occluded scenes (Sideboard), while the networks trained with simplified ones ($\beta=4$) perform better in general scenes.

Additionally, we remove the different model components to show their contributions to the estimation performance.  As shown in \cref{abla_tab}, OPENet without the EPI-Transformer or the refinement would slightly degraded. Meanwhile, OPENet trained on real-world datesets perform better than OPENet-fast obviously, which indicates that the feature fusion and refinement strategy is critical to our network. The reason might be that severe noise and uneven lighting in real world have a serious influence on occlusion pattern selection. And the EPI-Transfomer and refinement module make our model more robust to handle these complicated scenes.

Finally, we also analyze the impact of different loss items in \cref{total_loss}. Metrics reported show that all of the components are effective.

\section{Conclusions}
\noindent \textbf{Limitations and Future Work:} Despite the common issues caused by severe noise or non-Lambertian materials on real-world datasets, the assumptions for 1D occlusion patterns can not cover some extreme cases like very fine grids in front of an object or very small circular holes.
Designing more occlusion patterns may resolve this problem and we leave this as future work. We would also apply OPAL for unstructured light fields. Finally, we would leverage acceleration tools like TensorRT to further improve the overall efficiency of our method.

\noindent \textbf{Conclusion:} In this paper, we proposed OPENet, an unsupervised framework for light field disparity estimation. 
The occlusion pattern aware loss (OPAL) is designed to guarantee accuracy in the occluded areas, improve the generalization capacity, and the overall run-time efficiency. 
We introduce EPI-Transformer and a refinement module to further improve the performance. 
Experimental results demonstrate that we significantly reduce the computation overhead and achieve more robust performance when compared with state-of-the-art supervised methods. 
More importantly, our method can effectively avoid the domain shift effect when generalizing to real-world scenarios. We believe OPAL and OPENet re-defined the balance among accuracy, generalization and efficiency for light field disparity estimation.



%
%
\bibliographystyle{splncs04}
\bibliography{egbib}
\end{document}